\documentclass{ifacconf}

\usepackage{graphicx}      
\usepackage{natbib}        

\usepackage{epsfig} 
\usepackage{amsmath} 
\usepackage{amssymb}  

\usepackage{bm}
\usepackage{caption}
\usepackage{subcaption}
\usepackage{xcolor}
\usepackage{physics}

\begin{document}
\begin{frontmatter}

\title{Dual Arm Impact-Aware Grasping through Time-Invariant Reference Spreading Control} 

\thanks[footnoteinfo]{© 2023 the authors. This work has been accepted to IFAC for publication under a Creative Commons Licence CC-BY-NC-ND.}


\author[TU]{Jari J. van Steen} 
\author[TU]{Abdullah Coşgun} 
\author[TU]{Nathan van de Wouw}
\author[TU]{Alessandro Saccon}

\address[TU]{Eindhoven University of Technology, Department ME, DC group. Eindhoven, The Netherlands (e-mail: j.j.v.steen@tue.nl).}

\begin{abstract}                
	With the goal of increasing the speed and efficiency in robotic dual arm manipulation, a novel control approach is presented that utilizes intentional simultaneous impacts to rapidly grasp objects. 
	This approach uses the time-invariant reference spreading framework, in which partly-overlapping ante- and post-impact reference vector fields are used. These vector fields are coupled via the impact dynamics in proximity of the expected impact area, minimizing the otherwise large velocity errors after the impact and the corresponding large control efforts.
	A purely spatial task is introduced to strongly encourage the synchronization of impact times of the two arms.
	An interim-impact control phase provides robustness in the execution against the inevitable lack of exact impact simultaneity and the corresponding unreliable velocity error. 
	In this interim phase, a position feedback signal is derived from the ante-impact velocity reference, which is used to enforce sustained contact in all contact points without using velocity error feedback. 
	With an eye towards real-life implementation, the approach is formulated using a QP control framework, and is validated using numerical simulations on a realistic robot model with flexible joints and low-level torque control.
\end{abstract}


\end{frontmatter}

	\section{Introduction}\label{sec:introduction}
	\vspace*{-0.1cm}
	In the field of logistics, faster automated solutions are desired for tasks where humans currently excel in speed and reliability due to our ability to utilize impacts, such as those occurring in swift pick-and-place operations. 
	Robotic dual arm manipulation is worth exploring in palletizing and depalletizing applications, as it allows for a human-like pick-and-place motion without the need to create a custom end effector for different types of handled products \citep{Smith2012,Benali2018}. Utilization of impacts in dual arm manipulation in a human-like way can decrease cycle times, but is not a not an easily translatable skill. 
	Firstly, it needs to be ensured that hardware does not get damaged due to high peak forces \citep{Dehio2022}. 
	But even when impacts stay within safe limits, the rapid velocity transitions that result from impacts can result in peaks in the tracking error due to time or location mismatch between planned and actual collisions \citep{Biemond2013,Leine2008}. This, in turn, can generate unnecessarily large actuator commands, inducing extra vibrations and energy consumption.  
	
	This work focuses on the development of a framework for dual arm impact-aware manipulation under \textit{nominally simultaneous impacts} \citep{Rijnen2019}, meaning that the robot end-effectors' contact areas are ideally supposed to impact the object at the same time. 
	Misalignments and loss of impact simultaneity are however inevitable in practice, due to modeling and perception inaccuracies. 
	As a result, a series of unplanned intermediate impacts and a corresponding unpredictable series of velocity jumps will typically occur, where it is also not possible to estimate the contact state. As shown in \citet{Steen2022,Steen2022a}, this implies that velocity error feedback control cannot be reliably used during an impact sequence.
	
	In recent years, a handful of control methods have been developed to accurately execute motions that include simultaneous impacts while avoiding unwanted spikes in the control inputs. 
	This includes tracking approaches such as \citet{Yang2021,Morarescu2010}, and within the framework of reference spreading \citet{Rijnen2019,Steen2022,Steen2022a}. Reference spreading has been introduced in \citet{Saccon2014} and expanded in \citet{Rijnen2015,Rijnen2019a,Rijnen2020} for single-impact scenarios. It is a hybrid tracking control approach that deals with impacts by defining ante- and post-impact references that are coupled by an impact map at the nominal impact time and overlap around this time. It is ensured that the tracking reference corresponds with the actual contact state of the robot by switching the reference based on impact detection, which avoids error peaking and related spikes in the control inputs. In \citet{Rijnen2019}, a strategy to deal with simultaneous impacts is proposed, defining an interim-impact phase that uses only feedforward signals as long as contact is only partially established. This framework is extended in \citet{Steen2022} by also using position feedback during this interim phase to pursue persistent contact establishment without relying on velocity feedback. Furthermore, the approach in \citet{Steen2022} is cast into the quadratic programming (QP) robot control framework \citep{Bouyarmane2019,Salini2010}, similar to other impact-aware QP control approaches like \citet{Dehio2022, Wang2019}. This allows to include collision avoidance and joint limit constraints, which are essential in real-life robotic applications. 
	
	While \citet{Rijnen2019,Steen2022,Yang2021,Morarescu2010} use traditional \textit{time-based} references, \citet{Steen2022a} has introduced for the first time a \textit{time-invariant} version of reference spreading. As opposed to time-based tracking, the time-invariant nature of this approach enjoys the positive features of path following \citep{Aguiar2005} and maneuver regulation \citep{Hauser1995} approaches, preventing the systems to unnecessarily accelerate or decelerate in the presence of disturbances and deviations caused by temporary collision avoidance or conflicting tasks. 
	The approach also suits a large range of initial conditions without requiring replanning of the reference. 
	The reference in \citet{Steen2022a} is prescribed by desired anti- and post-impact vector fields, 	which overlap in position around the surface where the impact is expected to occur, to make sure a reference corresponding to the contact state is also at hand when the impact occurs away from this expected surface. At this surface, the ante- and post-impact velocity fields are coupled by an impact map \citep{Glocker2006,Brogliato2016}. 
	In \citet{Steen2022a}, the interim phase control developed in \citet{Steen2022} is extended to the case of time-invariant reference vector fields, constructing a position feedback signal through time integration of the ante-impact velocity reference. 
	
	Recent approaches, like \citet{Khurana2021} and \citet{Bombile2022}, focusing on learning to hit and dual arm grabbing, also employ such a time-invariant approach and originate from the well-known framework of dynamical systems-based robot control \citep{Salehian2018}. These approaches, however, focus on motion generation and not on control issues related to performing tasks with nominally simultaneous impacts and are thus complementary to the present work.  
	
	The main contribution of this paper is the formulation of a control approach for swift dual arm manipulation of objects through nominally simultaneous impacts based on time-invariant reference spreading \citep{Steen2022a}. 
	Compared to \citet{Steen2022a}, which focuses on single-arm manipulation, this work also enables synchronization between the robots, encouraging simultaneous grasping with both end effectors. This is especially useful when the arms starts from an initial condition which is far from being symmetric. 	
	Rather than modifying the references to enable simultaneous contact establishment as done in other multi-robot time-invariant approaches like   \citet{MirrazaviSalehian2018, Bombile2022}, we opt to enforce simultaneous contact establishment by adding a purely spatial synchronization task to the QP controller. This allows the robots to synchronize despite using two independent velocity references. 
	Another novelty is the method of coupling the ante-impact and post-impact reference vector fields. Whereas \citet{Steen2022a} uses an analytical impact map assuming a single simultaneous frictionless impact to predict the post-impact robot state, this approach cannot be followed here as the dual arm use case makes explicit use of contact friction to manipulate objects. We propose a novel simulation-based approach to predict the post-impact state that has the potential of being used when the impact map is not analytically known, lending itself to be used for even more complex real-world tasks. Finally, in the post-impact phase, a new method is proposed to move the object to a desired position through reference velocity fields and contact force regulation tasks without using the measured state of the object, as this is not assumed to be available in real-world applications.
	
	The outline of this paper is as follows. In Section \ref{sec:dynamics}, we provide equations of motion of the planar dual arm system used to demonstrate the proposed control approach. 
	Section \ref{sec:path} presents the approach to formulate the ante- and post-impact reference velocity fields. Section \ref{sec:control_approach} then describes the control framework consisting of the ante-impact, interim, and post-impact phases, used to track these references. 
	In Section \ref{sec:numerical_validation}, we show a validation of the approach against two baseline methods through realistic numerical simulations, before drawing conclusions in Section \ref{sec:conclusion}.

	\section{System dynamics}\label{sec:dynamics}
	
\begin{figure}
		\centering
		\includegraphics[width=\linewidth]{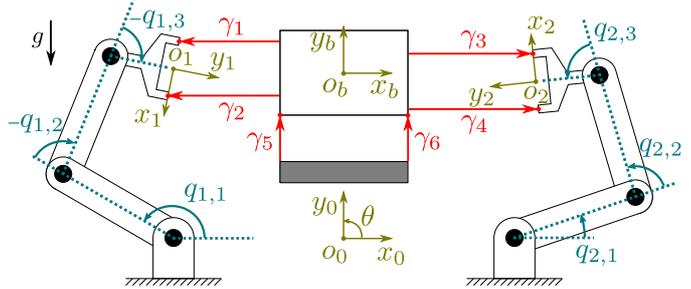}
		\caption{Overview of the two 3DOF robots impacting a box.}\label{fig:robots_3DOF}
	\end{figure}

	The control approach proposed in this paper is developed to cater towards the needs of real-life robotic applications. Without loss of generality and for illustration purposes, we consider the planar scene consisting of two 3DOF robots and a rigid box as shown in Figure \ref{fig:robots_3DOF} throughout this work. 
	
	For both robots, indicated by index $i \in \{1,2\}$, the position and orientation of the frame $o_ix_iy_i$ attached to the robot's end effector with respect to the fixed inertial frame $o_0x_0y_0$, is indicated by $\bm p_i$ and $\theta_i$, respectively. The position and orientation of the frame $o_bx_by_b$ attached to the center of the box with respect to $o_0x_0y_0$, are indicated by $\bm p_b$ and $\theta_b$. 
	The generalized coordinates are denoted by robot joint positions $\bm q_i$ as indicated in Figure \ref{fig:robots_3DOF}, and $\bm q_b = [\bm p_b, \theta_b]$. 
	The end effector and box velocities are denoted by
	\begin{equation}
	\begin{bmatrix}\dot{\bm p}_i \\ \dot{\theta}_i\end{bmatrix} = \begin{bmatrix}\bm J_{p,i}(\bm q_i) \\ \bm J_{\theta,i}(\bm q_i)\end{bmatrix} \dot{\bm{q}}_i, \ \ \ \begin{bmatrix}\dot{\bm p}_b \\ \dot{\theta}_b\end{bmatrix} = \begin{bmatrix}\bm J_{p,b}(\bm q_b) \\ \bm J_{\theta,b}(\bm q_b)\end{bmatrix} \dot{\bm{q}}_b
	\end{equation}
	with Jacobians
	\begin{equation}\label{eq:Jacobians}
	\begin{bmatrix}\bm J_{p,i}(\bm q_i) \\ \bm J_{\theta,i}(\bm q_i)\end{bmatrix} = \frac{\partial}{\partial \bm q_i}\begin{bmatrix}\bm p_i \\ \theta_i\end{bmatrix},	\ \ \	\begin{bmatrix}\bm J_{p,b}(\bm q_b) \\ \bm J_{\theta,b}(\bm q_b)\end{bmatrix} = \frac{\partial}{\partial \bm q_b}\begin{bmatrix}\bm p_b \\ \theta_b\end{bmatrix}.
	\end{equation}
	For ease of notation, in the following, we will drop the explicit dependency on $\bm q_i$, $\bm q_b$, $\dot{\bm q}_i$ and $\dot{\bm q}_b$. The equations of motion for the robots and box are, for $i \in \{1,2\}$, given by 
	\begin{align}\label{eq:EOM_robot}
	\bm M_i \ddot{\bm q}_i + \bm h_i &= \bm J_{N,i}^T  \bm \lambda_{N} + \bm J_{T,i}^T  \bm \lambda_{T} + \bm \tau_i,\\
	\bm M_b \ddot{\bm q}_b + \bm h_b &= \bm J_{N,b}^T  \bm \lambda_{N} + \bm J_{T,b}^T  \bm \lambda_{T},\label{eq:EOM_box}
	\end{align}
	with mass matrices $\bm M_i, \bm M_b \in \mathbb{R}^{3 \cross 3}$, gravitational, centrifugal and Coriolis terms $\bm h_i, \bm h_b \in \mathbb{R}^3$, applied joint torques $\bm \tau_i \in \mathbb{R}^3$, normal and tangential contact forces $\bm \lambda_{N},\bm \lambda_{T} \in \mathbb{R}^6$, and corresponding normal and tangential contact Jacobians $\bm J_{N,i},\bm J_{N,b},\bm J_{T,i},\bm J_{T,b} \in \mathbb{R}^{6 \cross 3}$. 
	Assuming no analytical impact map is at hand as addressed in Section \ref{sec:introduction}, the normal contact forces $\bm \lambda_{N}$ during and after the impact sequence are determined using a compliant contact model. For this, we used a contact model based on the exponentially extended Hunt-Crossley model (see (38) in \citet{Carvalho2019}) that avoids force jumps at contact establishment of the simpler Kelvin-Voigt model, as well as artificial adhesion during detachment of the classical Hunt-Crossley model. It is given by
	\begin{equation}\label{eq:contact_normal}
	\lambda_{N,k} = \left\{
	\begin{aligned}
	\mathcal{K}(\dot{\gamma}_k)(-\gamma_k)^c \quad & \text{if} \ \gamma_k \leq 0,  \\
	0 \quad & \text{if} \ \gamma_k > 0, 
	\end{aligned}
	\right.
	\end{equation}
	where negative $\gamma_k \in \mathbb{R}$, depicted in Figure \ref{fig:robots_3DOF}, denotes compenetration of the contacting surfaces, $c \in \mathbb{R}^+$ is a geometry dependent parameter, and
	\begin{equation}
	\mathcal{K}(\dot{\gamma}_k) = \left\{
	\begin{aligned}
	k_\gamma - d_\gamma\dot{\gamma}_k \quad & \text{if} \ \dot{\gamma}_k \leq 0,  \\
	k_\gamma \exp{\left(-\frac{d_\gamma}{k_\gamma}\dot{\gamma}_k\right)} \quad & \text{if} \ \dot{\gamma}_k > 0, 
	\end{aligned}
	\right.
	\end{equation}
	denotes the stiffness-damping factor, with stiffness and damping parameters $k_\gamma$ and  $d_\gamma$, respectively. The tangential contact force is determined through a smoothened Coulomb friction model \citep{Leine2004} as
	\begin{equation}
	\lambda_{T,k} = \mu \lambda_{N,k}\frac{2}{\pi}\arctan(\epsilon \dot{\sigma}_k)
	\end{equation}
	with $\dot{\sigma}_k$ as the relative tangential velocity perpendicular to $\dot{\gamma}_k$, friction coefficient $\mu \in [0,1]$ and shaping parameter $\epsilon \in \mathbb{R}^+$ determining the slope at zero tangential velocity.

	\section{Desired path generation}\label{sec:path}
	
	In this section, a possible formulation of a time-invariant velocity reference for the robots in the ante-impact phase is given, as well as a method to formulate a coupled velocity reference for the box in the post-impact phase. 

	\subsection{Ante-impact reference}\label{sec:path_ante}
	
	The ante-impact reference velocity field for each robot consists of an extended desired linear velocity $\bar{\dot{\bm p}}_{i,d}^a(\bm p_i)$ and a desired angular velocity ${\dot{\theta}}_{i,d}^a(\theta_i)$, with robot index $i\in \{1,2\}$. 
	Its goal is guiding the end effector to impact the box at a desired location with a desired speed.
	
	\subsubsection{Linear velocity reference}
	
\begin{figure}
		\centering
		\begin{subfigure}[b]{\linewidth}
			\centering
			\includegraphics[width=\textwidth]{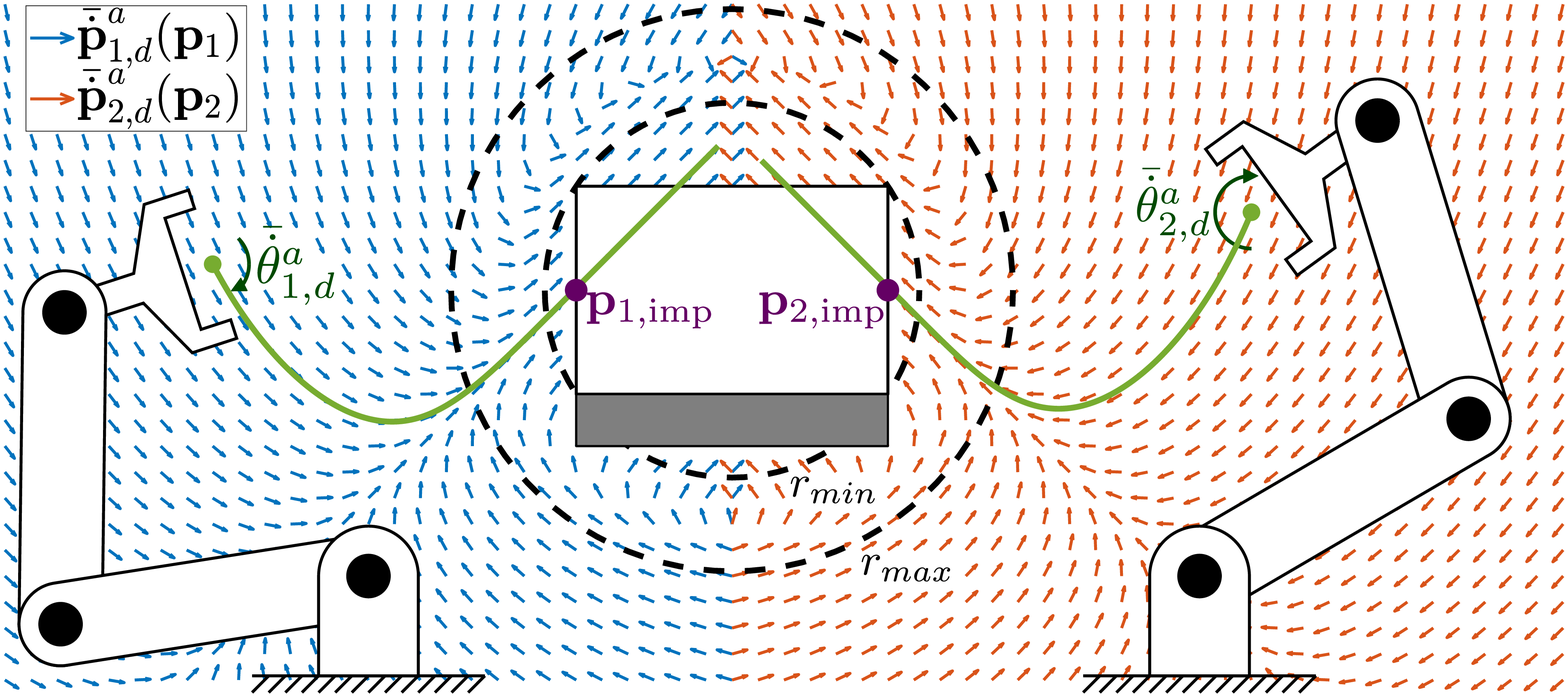}
			\caption{Ante-impact velocity reference with nominal end effector path.}
			\label{fig:ante_reference}
		\end{subfigure}
		\begin{subfigure}[b]{0.55\linewidth}
			\centering
			\includegraphics[width=\textwidth]{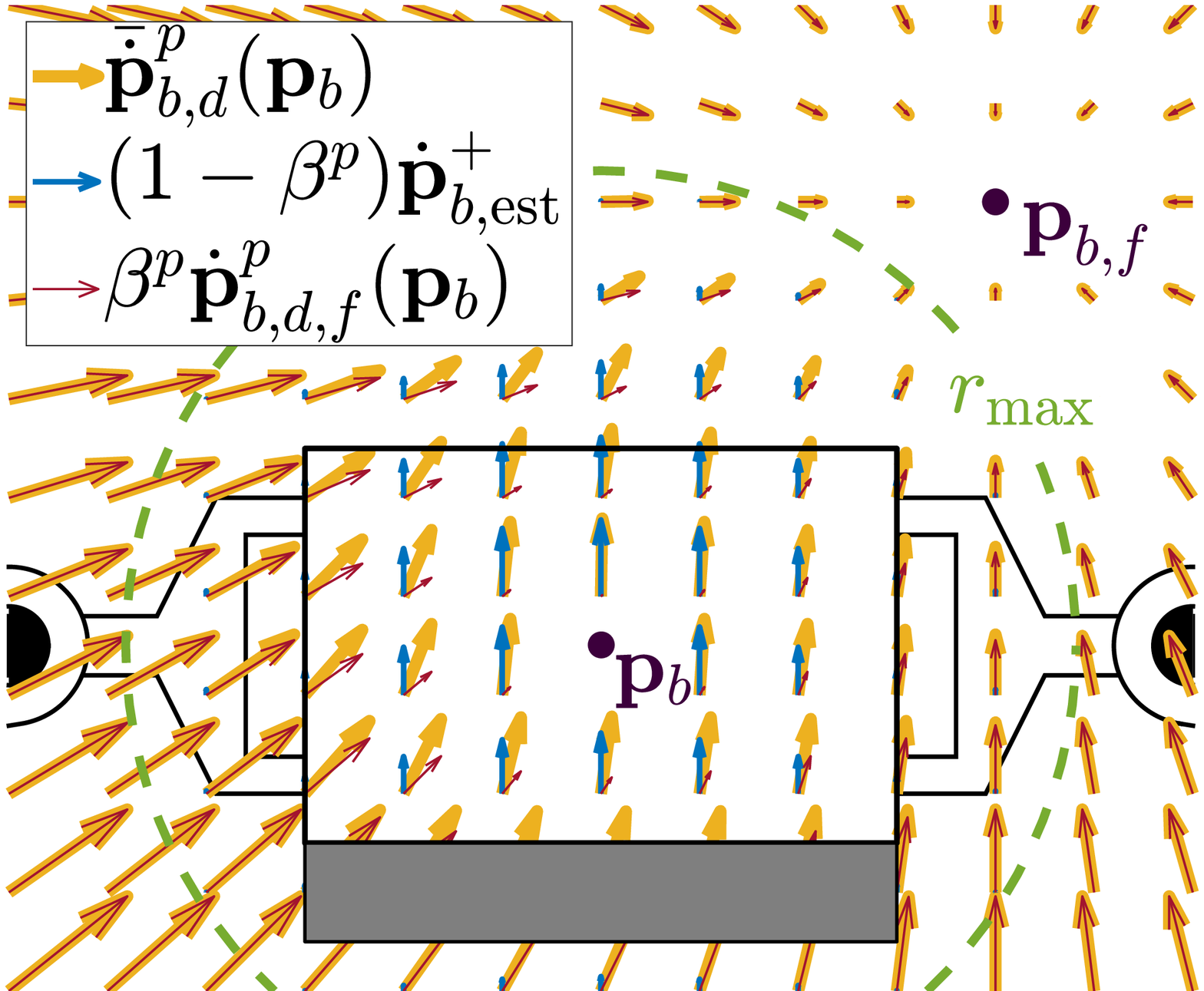}
			\caption{Post-impact velocity reference.}
			\label{fig:post_field}
		\end{subfigure}
		\hfill
		\begin{subfigure}[b]{0.34\linewidth}
			\centering
			\includegraphics[width=\textwidth]{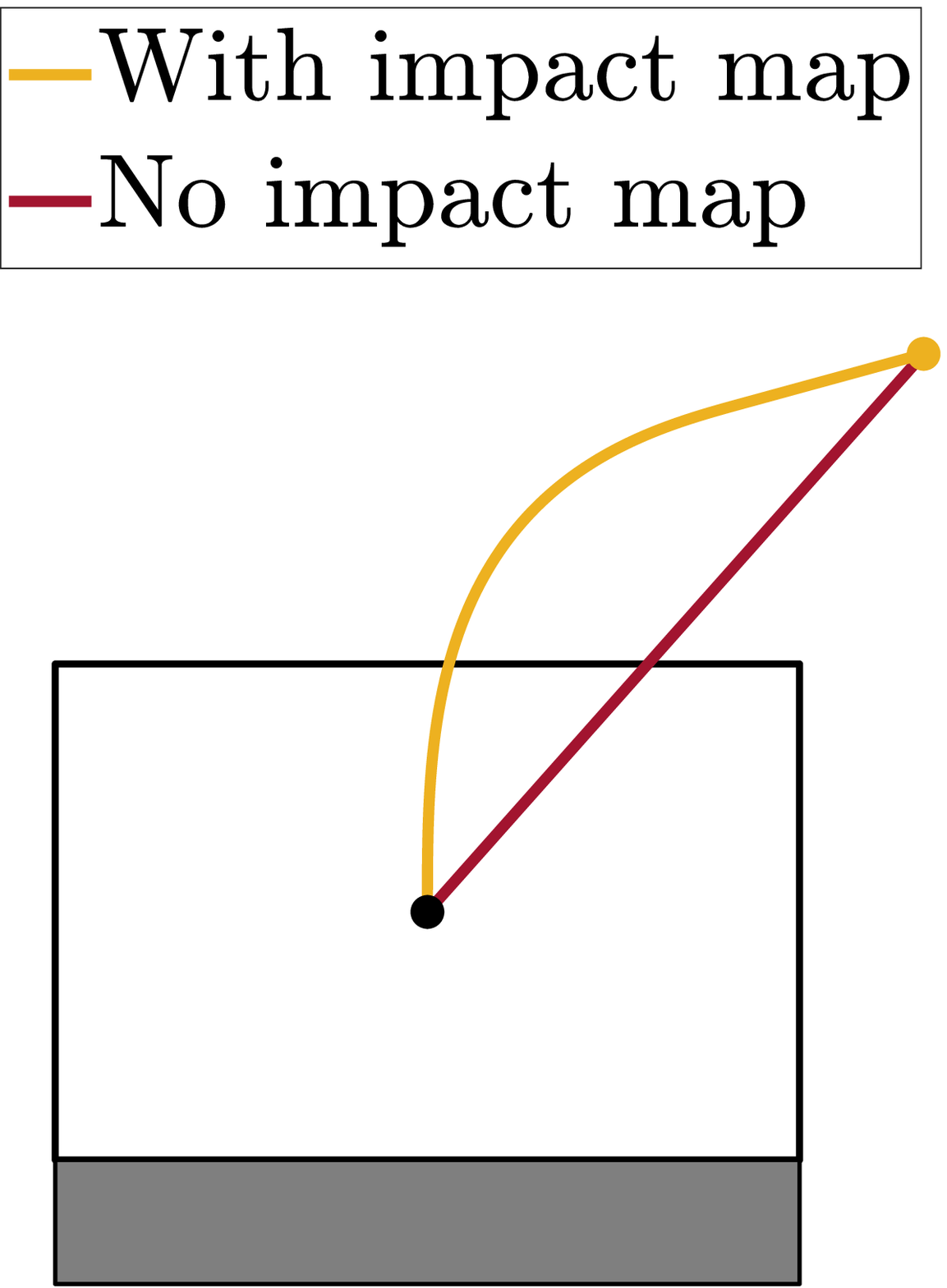}
			\caption{Nominal box path.}
			\label{fig:post_path}
		\end{subfigure}
		\caption{Reference fields for the ante- and post-impact motion.}
		\label{ch3: fig: post_reference_complete}
	\end{figure}

	To formulate the linear velocity reference for both end effectors, shown in Figure \ref{fig:ante_reference}, a desired impact location $\bm p_{i,\text{imp}}$ is defined together with a desired impact velocity vector $\bm v_{i,\text{imp}}$, with $\bm p_{i,\text{imp}}$ 
	taken as the estimated center of the box surface that robot $i$ is impacting. As in \citet{Steen2022a}, at first, an unextended velocity reference can be defined as
	\begin{equation}\label{eq:lin_ref_ante}
	{\dot{\bm p}}^a_{i,d}(\bm p_i) = \frac{\bm v_{i,\text{imp}} + \alpha(\bm p_{i,t}(\bm p_i) - \bm p_i)}{\norm{\bm v_{i,\text{imp}} + \alpha(\bm p_{i,t}(\bm p_i) - \bm p_i)}}\norm{\bm v_{i,\text{imp}}}
	\end{equation}
	with user-defined shaping parameter $\alpha \in \mathbb{R}^+$ and intermediate target position
	\begin{equation}\label{eq:p_t}
	\bm p_{i,t}(\bm p_i) = \bm p_{i,\text{imp}} -  \frac{\bm v_{i,\text{imp}}\norm{\bm p_{i,\text{imp}} - \bm p_i}}{\norm{\bm v_{i,\text{imp}}}}.
	\end{equation}
	This will result in a vector field that can guide each end effector towards $\bm p_{i,\text{imp}}$ until the end effector moves past $\bm p_{i,\text{imp}}$, at which point ${\dot{\bm p}}^a_{i,d}(\bm p_i)$ will guide the end effector back to $\bm p_{i,\text{imp}}$. 
	However, uncertainties in the location or dimensions of the box can cause the impact to occur past $\bm p_{i,\text{imp}}$. Hence, an extension of the reference is required to ensure that there is always a suitable ante-impact reference at hand that continues to guide the end effector towards the box rather than away from it. This extended velocity reference is depicted in Figure \ref{fig:ante_reference}, and is determined via
	\begin{equation}\label{eq:dp_d}
	\bar{\dot{\bm p}}_{i,d}^a(\bm p_i) = \beta^a_i {\dot{\bm p}}^a_{i,d}(\bm p_i) + (1-\beta^a_i)\bm v_{i,\text{imp}},
	\end{equation}
	with
	\begin{equation}
	\beta^a_i = S_1(\norm{\bm p_i - \bm{p}_{b,\text{est}}}),
	\end{equation}
	where $\bm{p}_{b,\text{est}}$ is the estimated center of the box, and $S_1: \mathbb{R} \to \mathbb{R}$ is the first-order smoothstep function 
	\begin{equation} \label{eq:smoothstep}
	S_1(r) = 
	\left\{\begin{aligned}
	0 & \text { if } r \leq r_{\mathrm{min}}, \\
	3 r_w^2 - 2 r_w^3  & \text { if } r_{\mathrm{min}} <  r < r_{\mathrm{max}}, \\
	1 & \text { if } r \geq r_{\mathrm{max}} \\
	\end{aligned}\right.
	\end{equation}
	with user-defined $r_{\mathrm{min}}, r_{\mathrm{max}} \in \mathbb{R}^+$ such that $r_{\mathrm{max}} > r_{\mathrm{min}} > \norm{\bm p_{i,\text{imp}} - \bm{p}_b}$ for $i \in \{1,2\}$, and
	\begin{equation}
	r_w = \frac{r - r_{\mathrm{min}}}{r_{\mathrm{max}} - r_{\mathrm{min}}}.
	\end{equation}
	This creates two circles around the center of the box with radii $r_{\mathrm{min}}$ and $r_{\mathrm{max}}$. The area outside of the larger circle corresponds to the vector field ${\dot{\bm p}}_{i,d}^a(\bm p_i)$ in \eqref{eq:lin_ref_ante} being followed, the area inside the smaller circle corresponds to a reference velocity $\bm v_{i,\text{imp}}$, and in between the two circles, a convex combination of ${\dot{\bm p}}_{i,d}^a(\bm p_i)$ and $\bm v_{i,\text{imp}}$ is taken, as depicted in Figure \ref{fig:ante_reference}. Since it holds that $r_{\mathrm{min}} > \norm{\bm p_{i,\text{imp}} - \bm{p}_b}$, it is ensured that, when moving past the desired impact location $\bm p_{i,\text{imp}}$, the reference velocity $\bm v_{i,\text{imp}}$ is prescribed, which continues to drive the robot towards the box. By using the smoothstep function, it is ensured that $\bar{\dot{\bm p}}_{i,d}^a(\bm p_i)$ is $C^1$ smooth, hence avoiding potential error spikes when following the reference.
	
	\subsubsection{Angular velocity reference}
	
	With the goal of aligning the robot's end effector surfaces with that of the box, a desired angular velocity is formulated as
	\begin{equation}\label{eq:dtheta_d}
	\dot{\theta}_{i,d}^a(\theta_{i}) = \kappa^a_{\theta}(\theta_{i,d} - \theta_{i})
	\end{equation}
	with user-defined $\kappa^a_{\theta} \in \mathbb{R}^+$, and $\theta_{i,d}$ for both robots as
	\begin{align}
	\theta_{1,d} & = \theta_{b,\text{est}} + 2 z_1 \pi,\\
	\theta_{2,d} & = \theta_{b,\text{est}} + \pi + 2 z_2 \pi
	\end{align}
	with estimated box orientation $\theta_{b,\text{est}}$ and $z_i \in \mathbb{Z}$ chosen such that $-\pi < \theta_{i,d} - \theta_{i} \leq \pi$. The reader should take \eqref{eq:dp_d} and \eqref{eq:dtheta_d} only as a possible choice, as illustration of the end-effector motion generation.

	\subsection{Post-impact reference}\label{sec:path_post}
	
	Given the fact that the robots and box are assumed to be in contact in the post-impact phase, a linear and angular velocity reference is prescribed for the box, denoted by $\bar{\dot{\bm p}}_{b,d}^p(\bm p_b)$ and $\bar{\dot{\theta}}_{b,d}^p(\theta_b)$, respectively. Without loss of generality, we assume that the goal of the post-impact reference is to guide the box to a desired final pose denoted by $(\bm p_{b,f},\theta_{b,f})$. Still, as described in Section \ref{sec:introduction}, we aim to do so while ensuring that the ante- and post-impact references are compatible with the dynamics of the system, minimizing velocity errors and input spikes after contact is established. So in order to determine $\bar{\dot{\bm p}}_{b,d}^p(\bm p_b)$ and $\bar{\dot{\theta}}_{b,d}^p(\theta_b)$, the post-impact velocity is predicted based on the state of the system when entering the post-impact phase and the ante-impact velocity reference.
	
	\subsubsection{Post-impact nominal velocity prediction}\label{sec:post_vel_pred}
	
	Instead of the analytical approach taken in \citet{Steen2022a},  
	we explore here the alternative option of predicting the post-impact velocity by means of (offline) numerical simulations. This approach is particularly useful when an analytical impact map cannot be determined. 
	In these simulations, different simultaneous impact configurations are sampled given the estimated ante-impact pose of the box $(\bm p_{b,\text{est}}, \theta_{b,\text{est}})$. The end effector velocities for both robots are initialized to match the desired velocity determined in Section \ref{sec:path_ante}. The post-impact box velocity for each sampled configuration is determined through these simulations and, through radial basis function (RBF) interpolation, the nominal post-impact linear and angular box velocity $\dot{\bm p}^+_{b,\text{est}}$ and $\dot{\theta}^+_{b,\text{est}}$ is determined for the actual impact configuration. Further details can be found in Appendix \ref{app:path_post}.
	
	

	\subsubsection{Linear velocity reference}
	
	With the goal of formulating a vector field that drives the box towards a final position $\bm p_{b,f}$ while also assuring consistency with the ante-impact reference, 
	the desired velocity $\bar{\dot{\bm p}}_{b,d}^p(\bm p_b)$ is constituted of two terms. The first term is $\dot{\bm p}^p_{b,d,f}$, which drives the box to a desired final position $\bm p_{b,f}$, and is given by
	\begin{equation} \label{eq:attractor}
	\dot{\bm p}^p_{b,d,f}(\bm p_b) = \kappa^p_p (\bm p_{b,f} - \bm p_{b})
	\end{equation}
	with user-defined $\kappa^p_p \in \mathbb{R}^+$. 
	The second term is given by the nominal post-impact box velocity $\dot{\bm p}^+_{b,\text{est}}$, as obtained through the approach detailed earlier in this section. The two are blended to obtain the extended velocity reference
	\begin{equation} \label{eq:ante_ext_vel}
	\bar{\dot{\bm p}}_{b,d}^{p}(\bm p_{b}) = \beta^{p}\dot{\bm p}_{b,d,f}^{p}(\bm p_{b}) + (1-\beta^{p})\dot{\bm p}^+_{b,\text{est}},
	\end{equation}
	with
	\begin{equation} \label{eq:post_transition}
	\beta^{p} = S_1\left(||{\bm p^+_{b,est} - \bm p_b}||\right)
	\end{equation}
	and smoothstep $S_1(r)$ according to \eqref{eq:smoothstep} with $r_\text{min}=0$, $0 < r_\text{max} \leq ||\bm p_{b,f} - \bm p^+_{b,est}||$,
	and $\bm p^+_{b,est}$ as the estimated box position just after the impact. This combination is visualized in Figure \ref{fig:post_field}. 
	A comparison between the nominal box path according to the proposed approach, or if the impact is not accounted for, is given in Figure \ref{fig:post_path}. From this figure, it becomes apparent that, given the vertical box velocity directly after the impact, the nominal box path with impact map is the more natural one that, when followed, will avoid a large jump in the input torques.
	
	\subsubsection{Angular velocity reference}
	
	The desired angular box velocity is formulated using the terms $\dot{\theta}^+_{b,\text{est}}$ and $\dot{\theta}_{b,d,f}^{p}$, with
	\begin{equation} \label{eq:attractor_th}
	\dot{\theta}^p_{b,d,f}(\theta_b) = \kappa^p_\theta (\theta_{b,f} - \theta_{b})
	\end{equation}
	for user-defined $\kappa^p_\theta \in \mathbb{R}^+$ and desired final orientation $\theta_{b,f}$. Similar to the linear reference velocity formulation, the extended desired angular velocity is given by
	\begin{equation} \label{eq:ante_ext_rot}
	\bar{\dot{\theta}}_{b,d}^{p}(\theta_{b}) = \beta^{p}\dot{\theta}_{b,d,f}^{p}(\theta_{b}) + (1 - \beta^{p})\dot{\theta}^+_{b,\text{est}}
	\end{equation}
	with $\beta^p$ determined by \eqref{eq:post_transition}, causing also the angular velocity reference to be compatible with the ante-impact reference, while still guiding the box towards its desired orientation.

	\section{Control Approach}\label{sec:control_approach}
	
	Using the velocity references as defined in Section \ref{sec:path}, a control input is constructed for the ante-impact, interim, and post-impact phase using the time-invariant reference spreading framework. For each of the three phases, a QP controller is designed. 
	The switching policy between the phases is based on detection of the first impact of either robot, which activates the interim phase, and then monitoring when full contact is established for both end effectors, which activates the post-impact phase. Each of the phases has a corresponding discrete-time QP controller which is used to obtain a desired joint torque $\bm \tau^*$ to be applied to the robot after every fixed time step $\Delta t$. 
	
	Compared to \citet{Steen2022a}, which introduced time-invariant reference spreading for a single robot arm use case, simply following the ante-impact velocity reference for both robots might result in a large mismatch in impact timing between the left and right arm, 
	which is to be accounted for in the ante-impact phase. Furthermore, the interim and post-impact phases are adjusted for the new dual arm use case, where for the post-impact phase, force tasks are formulated that regulate contact forces, while the box is guided towards its final position without relying on any position or velocity measurements of the box, as this information will likely also not be available in real-life applications. 
	
	\subsection{Ante-impact phase}\label{sec:control_ante}
	
	In the ante-impact phase, the QP optimization variables are given by the input torques $\bm \tau_1$ and $\bm \tau_2$ in \eqref{eq:EOM_robot} that we seek to obtain and the joint accelerations $\ddot{\bm q}_1$ and $\ddot{\bm q}_2$. 
	The ante-impact QP controller is designed with five tasks, one task to enforce synchronization between the end effector positions relative to the estimated box position, two tasks to track both end effector's linear velocities, and two tasks to track both end effector's angular velocities. 
	A weighted cost function is formulated creating a weak hierarchy between the tasks, with the intent of realizing almost simultaneous impacts. 
	
	To encourage synchronization between the end effectors with respect to the box, we propose to mirror $\bm p_{2}$ along the axis $y_B$ of the box frame such that it can be compared to position $\bm p_1$. This mirrored position $\bm p_{2,m}$ is given by
	\begin{equation} \label{eq:sync_point}
	\bm p_{2,m} = \bm T_{s}(\bm p_{2} - \bm p_{b,\text{est}}) + \bm p_{b,\text{est}}
	\end{equation}
	with 
	\begin{equation}
	\bm T_s = \begin{bmatrix}
	c_{\theta_{b,\text{est}}} & -s_{\theta_{b,\text{est}}} \\
	s_{\theta_{b,\text{est}}} & c_{\theta_{b,\text{est}}}
	\end{bmatrix}\begin{bmatrix}
	-1 & 0 \\ 0 & 1
	\end{bmatrix}\begin{bmatrix}
	c_{\theta_{b,\text{est}}} & -s_{\theta_{b,\text{est}}} \\
	s_{\theta_{b,\text{est}}} & c_{\theta_{b,\text{est}}}
	\end{bmatrix}^T.
	\end{equation}
	Assuming the box is stationary in the ante-impact phase, the mirrored velocity and acceleration are given by
	\begin{equation*} \label{eq:sync_vel}
	\dot{\bm p}_{2,m} = \bm T_{s}\bm J_{p,2}\dot{\bm q}_2, \ \ \ \ddot{\bm p}_{2,m} = \bm T_{s}(\bm J_{p,2}\ddot{\bm q}_2 + \dot{\bm J}_{p,2}\dot{\bm q}_2).
	\end{equation*}
	Using this definition, the error $\bm e^a_{s}$ can be defined as
	\begin{equation}
	\bm e^a_{s} := \ddot{\bm p}_{1}-\ddot{\bm p}_{2,m} - k^\text{sync}_{p,p}({\bm p}_{2,m}-{\bm p}_{1}) - k^\text{sync}_{p,d}(\dot{\bm p}_{2,m}-\dot{\bm p}_{1})
	\end{equation}
	with user-defined gains $k^\text{sync}_{p,d}, k^\text{sync}_{p,p} \in \mathbb{R}^+$. 
	The error can be rewritten in terms of $\ddot{\bm q}_1$ and $\ddot{\bm q}_2$, as
	\begin{equation}
	\begin{aligned}
	& \bm e^a_{s} = \bm J_{p,1}\ddot{\bm q}_1 + \dot{\bm J}_{p,1}\dot{\bm q}_1 - \bm T_s(\bm J_{p,2}\ddot{\bm q}_2 + \dot{\bm J}_{p,2}\dot{\bm q}_2)  \\ & - k^\text{sync}_{p,p}({\bm p}_{2,m}-{\bm p}_{1}) - k^\text{sync}_{p,d}({\bm T_{s}{\bm J}_{p,2}\dot{\bm q}_2-\bm J}_{p,1}\dot{\bm q}_1).
	\end{aligned}
	\end{equation}
	Similarly to \citet{Steen2022a}, the error related to tracking the linear velocity reference for robot $i$ is 
	\begin{equation}\label{eq:pos_task_ante}
	\bm e^a_{p,i} = \bm J_{p,i}\ddot{\bm q}_i + \dot{\bm J}_{p,i}\dot{\bm q}_i - \bar{\ddot{{\bm p}}}^a_{i,d}(\bm p_i) - k^a_p \left(\bar{\dot{{\bm p}}}^a_{i,d}(\bm p_i) - \bm J_{p,i}\dot{\bm q}_i\right)
	\end{equation}
	with user-defined gain $k^a_p \in \mathbb{R}^+$, and desired acceleration $\bar{\ddot{\bm p}}^a_{i,d}(\bm p_i)$ derived from $\bar{\dot{{\bm p}}}^a_{i,d}(\bm p_i)$ through 
	\begin{equation}\label{eq:ddot_p_a_d}
	\bar{\ddot{\bm p}}^a_{i,d}(\bm p_i) = \frac{\partial \bar{\dot{{\bm p}}}^a_{i,d}}{\partial \bm p_i}(\bm p_i) \bar{\dot{{\bm p}}}^a_{i,d}(\bm p_i).
	\end{equation}
	The angular velocity tracking error is given by
	\begin{equation}\label{eq:rot_task_ante}
	e^a_{\theta,i} = \bm J_{\theta,i}\ddot{\bm q}_i + \dot{\bm J}_{\theta,i}\dot{\bm q}_i - {\ddot{{\theta}}}^a_{i,d}(\theta_i) - k^a_\theta ({\dot{{\theta}}}^a_{i,d}(\theta_i) - \bm J_{\theta,i}\dot{\bm q}_i)
	\end{equation}
	with user-defined gain $k^a_\theta \in \mathbb{R}^+$, and desired acceleration $\bar{\ddot{\theta}}^a_{i,d}(\theta_i)$ derived in similar fashion to \eqref{eq:ddot_p_a_d}. The QP control cost function is then given by the weighted sum 
	\begin{equation}
	E_\text{ante} = \sum_{i=1}^2 \left(w^a_p\norm{\bm e^a_{p,i}}^2 + w^a_\theta \left|e^a_{\theta,i}\right|^2\right) + w^a_s \norm{\bm e^a_{s}}^2
	\end{equation}
	with user-defined weights $w^a_p, w^a_\theta,w^a_s \in \mathbb{R}^+$.
	Combining this cost function with the equations of motion of \eqref{eq:EOM_robot} with $\bm \lambda_N = \bm \lambda_T = 0$, and a constraint forcing the input torque to remain between lower and higher bounds $\underline{\bm \tau}$ and $\bar{\bm \tau}$, gives the full ante-impact QP
	\begin{equation}
	(\ddot{\bm q}_1^*, \ddot{\bm q}_2^*, \bm \tau_1^*, \bm \tau_2^*) = \underset{\ddot{\bm q}_1, \ddot{\bm q}_2,  \bm \tau_1, \bm \tau_2}{\operatorname{argmin}} \ E_\text{ante},
	\end{equation}
	s.t., for $i\in\{1,2\}$,
	\begin{align}\label{eq:EOM_ante}
	\bm M_i\ddot{\bm q}_i + & \ \bm h_i = \bm \tau_i, \\
	\underline{\bm \tau} \leq & \ \bm \tau_i \leq \bm \bar{\bm \tau}.\label{eq:torque_const}
	\end{align}
	The reference torques $\bm \tau_1^*$ and $\bm \tau_2^*$ are then obtained by solving this QP at all times $t_k$ separated by time step $\Delta t$, with $\bm q_i = \bm q_i(t_k)$, $\dot{\bm q}_i = \dot{\bm q}_i(t_k)$.
	
	\subsection{Interim phase}\label{sec:control_interim}
	
	As advocated in Section \ref{sec:introduction}, we cannot rely on velocity feedback control using either velocity reference since the system is neither in the ante- nor the post-impact state for which the references are defined. We also assume not to have contact force estimates during the interim phase, as the exact contact state cannot be reliably estimated in reality. The challenge of this interim phase is hence to formulate a QP tasked with establishing full sustained contact without relying on velocity error feedback. 
	Our approach aims to achieve this in similar fashion to \citet{Steen2022a} by initially applying torque as if the system is still in the ante-impact phase. We then add position feedback based on a position error constructed online using the ante-impact velocity reference and the pose of the robots at the beginning of the interim phase. This position reference can be iteratively determined by 
	\begin{equation}\label{eq:pos_int}
	\bm p_{i,d}^\text{int}(t_k + \Delta t) = \bm p_{i,d}^\text{int}(t_k) + \bar{\dot{\bm p}}^a_{i,d}(\bm p_{i,d}^\text{int}(t_k)) \Delta t
	\end{equation}
	with QP time step $\Delta t$ and $\bm p_{i,d}^\text{int}(t_\text{int}) = \bm p_i(t_\text{int})$, with $t_\text{int}$ as the time the interim phase is entered. 
	We then replace $\dot{\bm q}_i$ in \eqref{eq:pos_task_ante} by the nominal joint velocity ${\dot{\bm q}}_{i,\text{int}}$, given by
	\begin{equation}\label{eq:q_itmd}
	{\dot{\bm q}}_{i,\text{int}} = \begin{bmatrix}\bm J_{p,i} \\ \bm J_{\theta,i}\end{bmatrix}^{-1}  \begin{bmatrix} \bar{\dot{\bm p}}^a_{i,d}(\bm p_i) \\ \dot{{\theta}}^a_{i,d}(\theta_i)\end{bmatrix},
	\end{equation} 
	and add position feedback based on \eqref{eq:pos_int} to obtain the new interim cost
	\begin{equation}\label{eq:pos_task_int}
	{\bm e}^\text{int}_{p,i} = \bm J_{p,i}\ddot{\bm q}_i + {\dot{\bm J}}_{p,i,\text{int}}{\dot{\bm q}_{i,\text{int}}} - \bar{\ddot{\bm p}}^a_{i,d}(\bm p_i) - k^\text{int}_{p} \left({\bm p}^\text{int}_{i,d}(t) - {\bm p_i}\right)
	\end{equation}
	with
	\begin{equation}\label{eq:J_p_int}
	{\dot{\bm J}}_{p,i,\text{int}} := \sum_{i=1}^{3}\frac{\partial \bm J_{p,i}}{\partial q_i}{\dot{q}}_{i,\text{int}},
	\end{equation}
	and user-defined gain $k_p^\text{int} \in \mathbb{R}^+$. Due to the choice of ${\dot{\bm q}}_{i,\text{int}}$ in \eqref{eq:q_itmd}, the velocity feedback term of \eqref{eq:pos_task_ante} reduces to $\bar{\dot{{\bm p}}}^a_{i,d}(\bm p_i) - \bm J_{p,i} \dot{\bm q}_{i,\text{int}} = \bm 0$, which is why indeed no velocity feedback term is present in \eqref{eq:pos_task_int}. Similarly, the cost corresponding to the orientation task can be formulated as
	\begin{equation}\label{eq:ori_task_int}
	{\bm e}^\text{int}_{\theta,i} = \bm J_{\theta,i}\ddot{\bm q}_i + {\dot{\bm J}}_{\theta,i,\text{int}}{\dot{\bm q}_{i,\text{int}}} - \bar{\ddot{\theta}}^a_{i,d}(\theta_i) - k^\text{int}_{\theta} \left({\theta}^\text{int}_{i,d}(t) - {\theta_i}\right)
	\end{equation}
	with $k_p^\text{int} \in \mathbb{R}^+$ and ${\dot{\bm J}}_{\theta,i,\text{int}}$ defined as in \eqref{eq:J_p_int}. Combining both errors leads to the interim phase cost function
	\begin{equation}\label{eq:cost_int}
	E_\text{int} = \sum_{i=1}^2 \left(w^a_p\norm{\bm e^\text{int}_{p,i}}^2 + w^a_\theta \left|e^\text{int}_{\theta,i}\right|^2\right).
	\end{equation}
	No synchronization error needs to be added, as it is assumed that the end effectors have sufficiently synchronized before the interim phase is entered. 
	Regarding constraints, the equation of motion constraint \eqref{eq:EOM_ante} is modified, also replacing ${\dot{\bm q}}_i$ by ${\dot{\bm q}}_{i,\text{int}}$, resulting in the interim phase QP 
	\begin{equation}
	(\ddot{\bm q}_1^*, \ddot{\bm q}_2^*, \bm \tau_1^*, \bm \tau_2^*) = \underset{\ddot{\bm q}_1, \ddot{\bm q}_2,  \bm \tau_1, \bm \tau_2}{\operatorname{argmin}}	\ E_\text{int},
	\end{equation}
	s.t., for $i \in \{1,2\}$,
	\begin{align}\label{eq:EOM_interim}
	\bm M_i\ddot{\bm q}_i +  \bm h_i & (\bm q_i, \bm q_{i,\text{int}}) = \bm \tau_i, \\
	\underline{\bm \tau} \leq & \ \bm \tau_i \leq \bm \bar{\bm \tau}.
	\end{align}
	As a result of the continuity in the feedforward acceleration $\bar{\ddot{\bm p}}_{i,d}^a$ and $\ddot{\theta}_{i,d}^a$ between the ante-impact and interim mode, only a minimal input torque jump will occur when the interim mode is entered. While initially 0, the accumulation of the position feedback error over time results in a driving force to establish full contact with both robots until this is achieved. We assume stabilization in this phase is provided by the dissipating contact dynamics of \eqref{eq:contact_normal}.

	\subsection{Post-impact phase}\label{sec:control_post}
	
	After full sustained contact between both end effectors and the box is established, the post-impact phase is entered. This post-impact QP explicitly accounts for the contact forces, and hence, $\bm \lambda_N$ and $\bm \lambda_T$ are included in the optimization variables on top of $\ddot{\bm q}_1$, $\ddot{\bm q}_2$, $\bm \tau_1$, $\bm \tau_2$ and $\ddot{\bm q}_b$. 
	
	First of all, the formulation of the constraints is given. This includes the robot and box equations of motion of \eqref{eq:EOM_robot} and \eqref{eq:EOM_box}, as well as the input torque limit constraint from \eqref{eq:torque_const}. To ensure contact is not lost, the normal contact forces are constrained to be nonnegative, resulting in 
	\begin{equation}\label{eq:const_lambda_N}
	\bm \lambda_N > \bm 0.
	\end{equation}
	Furthermore, the contacts should stay within the friction cone, resulting in the constraint
	\begin{equation}\label{eq:const_lambda_T}
	-\mu_\text{est} \bm \lambda_N < \bm \lambda_T < \mu_\text{est} \bm \lambda_N,
	\end{equation}
	where $\mu_\text{est} \in [0,1]$ is an approximation of the friction coefficient for the robot-box contacts. As the solutions that result from this QP are likely to be on the boundary of the feasible domain, $\mu_\text{est} < \mu$ should hold to avoid slippage. With these constraints enforced, it is assumed that no tangential or normal acceleration takes place between the contacts, resulting in constraints $\ddot{\gamma}_k = 0$ for $k \in \{1,2,3,4\}$ and $\ddot{\sigma}_k = 0$ for $k \in \{1,3\}$. Note that $\ddot{\sigma}_2 = \ddot{\sigma}_4 = 0$ follows from these constraints, and is not additionally prescribed to avoid overconstraining the QP. In terms of the decision variables, the constraints are given by
	\begin{equation}\label{eq:normal_pos_const}
	\sum_{i=1}^2\left(\bm {J}_{N,i}\ddot{\bm q}_i + \dot{\bm {J}}_{N,i}\dot{\bm q}_i\right) - \bm {J}_{N,b}\ddot{\bm q}_b - \dot{\bm {J}}_{N,b}\dot{\bm q}_b = \bm 0,
	\end{equation}
	\begin{equation}\label{eq:tang_pos_const}
	\ddot{\sigma}_k\left(\ddot{\bm q}_1, \ddot{\bm q}_2, \ddot{\bm q}_b, \dot{\bm q}_1, \dot{\bm q}_2, \dot{\bm q}_b, \bm q_1, \bm q_2, \bm q_b\right) = \bm 0 \ \text{for} \ k \in \{1,3\}.
	\end{equation}
	Regarding tasks, the QP consists of a task that prescribes the linear velocity of the box, and one that prescribes its angular velocity, both according to the reference defined in Section \ref{sec:path_post}. The linear and angular velocity errors are similar in structure to \eqref{eq:pos_task_ante} and \eqref{eq:rot_task_ante}, resulting in
	\begin{equation}\label{eq:pos_task_post}
	\bm e^p_{p,b} = {\bm J}_{p,b}\ddot{\bm q}_b - \bar{\ddot{\bm p}}^p_{b,d}(\bm p_b) - k^p_p \left(\bar{\dot{\bm p}}^p_{b,d}(\bm p_b) - \bm J_{p,b}\dot{\bm q}_b\right),
	\end{equation}
	\begin{equation}\label{eq:rot_task_post}
	\bm e^p_{\theta,b} = {\bm J}_{\theta,b}\ddot{\bm q}_b - \bar{\ddot{\theta}}^p_{b,d}(\theta_b) - k^p_\theta \left(\bar{\dot{\theta}}^p_{b,d}(\theta_b) - \bm J_{\theta,b}\dot{\bm q}_b\right)
	\end{equation}
	with $k^p_p,k^p_\theta \in \mathbb{R}^+$, and $\bar{\ddot{\bm p}}^p_{b,d}(\bm p)$ and $\bar{\ddot{\theta}}^p_{b,d}(\theta)$ derived similarly to \eqref{eq:ddot_p_a_d}. Note that $\dot{\bm J}_{p,b}=\dot{\bm J}_{\theta,b}=\bm 0$ according to the definition in \eqref{eq:Jacobians}.
	As in \citet{Steen2022a}, we add a task encouraging an equal normal contact force distribution over both contact points of the end effector for each robot, resulting in two errors
	\begin{equation}\label{eq:force_dist}
	e_{\lambda,1} = \lambda_{N,1} - \lambda_{N,2}, \ \ \  e_{\lambda,2} = \lambda_{N,3} - \lambda_{N,4}.
	\end{equation}
	Furthermore, an error is formulated to minimize the norm of the normal contact forces within the feasible set defined by the constraints, leading to the QP cost function
	\begin{equation}\label{eq:cost_post}
	E_\text{post} = w^p_p\norm{\bm e^p_{p,b}}^2 + w^p_\theta \left|e^p_{\theta,b}\right|^2 + w^p_\lambda\left(e_{\lambda,1}^2 + e_{\lambda,2}^2\right) + w^p_{n} \norm{\bm \lambda_N}^2
	\end{equation}
	with task weights $w^p_p, w^p_\theta, w^p_\lambda, w^p_n \in \mathbb{R}^+$. 
	This leads to the full post-impact QP formulation as
	\begin{equation}
	(\ddot{\bm q}_1^*, \ddot{\bm q}_2^*, \ddot{\bm q}_b^*, \bm \tau_1^*, \bm \tau_2^*, \bm \lambda_N^*, \bm \lambda_T^*) =  \underset{\ddot{\bm q}_1, \ddot{\bm q}_2, \ddot{\bm q}_b, \bm \tau_1, \bm \tau_2, \bm \lambda_N, \bm \lambda_T}{\operatorname{argmin}} \	E_\text{post},
	\end{equation}
	s.t., for $i \in \{1,2\}$,
	
	\begin{center}
		\eqref{eq:EOM_robot}, \eqref{eq:EOM_box}, \eqref{eq:torque_const}, \eqref{eq:const_lambda_N}, \eqref{eq:const_lambda_T}, \eqref{eq:normal_pos_const}, \eqref{eq:tang_pos_const}.
	\end{center}
	
	As it is assumed that the box state cannot be measured accurately unlike the state of the robots, an estimation of the box state is used in the QP. This estimation assumes the robots grasp on average the center of the respective box surface, resulting in $\bm q_b$ and $\dot{\bm q}_b$ to be replaced with
	\begin{equation}
	\bm q_{b,\text{est}} = \begin{bmatrix}
	(\bm p_1 + \bm p_2)/2 \\ (\theta_1 + \theta_2)/2
	\end{bmatrix}, \ \  \dot{\bm q}_{b,\text{est}} = \begin{bmatrix}
	(\dot{\bm p}_1 + \dot{\bm p}_2)/2 \\ (\dot{\theta}_1 + \dot{\theta}_2)/2
	\end{bmatrix}.
	\end{equation}
	
	\section{Numerical validation}\label{sec:numerical_validation}
	
	To validate the proposed control approach, numerical simulations have been performed\footnote{All simulations can be reproduced using the publicly available Matlab scripts that can be found in https://gitlab.tue.nl/robotics-lab-public/dual-arm-impact-aware-grasping-through-time-invariant-reference-spreading-control.} tasking the robot to grasp the box through a simultaneous impact. The goal is to show the robustness of the proposed control approach against uncertainties in the environment, resulting in separated intermediate impacts instead of the planned simultaneous impact. 
	Despite the corresponding unpredictable series of velocity jumps and unknown contact states, it is shown that using the proposed control approach leads to a robust execution of the desired motion without unwanted spikes in the input torque. 
	
	We mimic the aforementioned uncertainty in the environment in the simulations by taking a wrong initial estimation of the box pose $q_{b,\text{est}}$, rotating the box by $2.5^{\circ}$ and moving the box somewhat in vertical direction. 
	Additionally, simulations are performed using a realistic robot model with flexibility modeled in its joints, combined with a low-level torque control law (see (7) in \citet{Albu2007}), that takes as input the desired joint torques computed through the proposed control approach. This flexible model more closely resembles reality, as flexibility is generally present in, e.g., the drivetrain of robot joints, resulting in oscillations as a result of impacts. This serves as validation that the developed approach, which uses the assumption of a rigid robot as explained in Section \ref{sec:dynamics}, is suitable for real-life robot control. 
	The system is also initialized in a state where the end effector pose of robot 1 differs significantly from that of robot 2 when mirrored along the axis $y_B$, to indicate that the time-invariant approach is robust to asymmetric initial conditions, with the synchronization task in the ante-impact QP eventually leading to a nearly simultaneous impact between both robots and the box. 
	
	To further show the benefit of the proposed control approach, simulations are also performed for two baseline approaches. In the first, referred to as the approach with \textit{no impact map}, the post-impact velocity prediction is not used to formulate the post-impact reference, implying that $\bar{\dot{\bm p}}_{b,d}^{p}(\bm p_{b})$ in the post-impact QP is substituted by ${\dot{\bm p}}_{b,d,f}^{p}(\bm p_{b})$ from \eqref{eq:attractor}. In the second baseline approach, referred to as the approach with \textit{no interim phase}, the control phase is switched from ante-impact to post-impact after full contact is established, hence staying in the ante-impact phase when contact is partially established.

\begin{figure}
		\centering
		\includegraphics[width=0.98\linewidth]{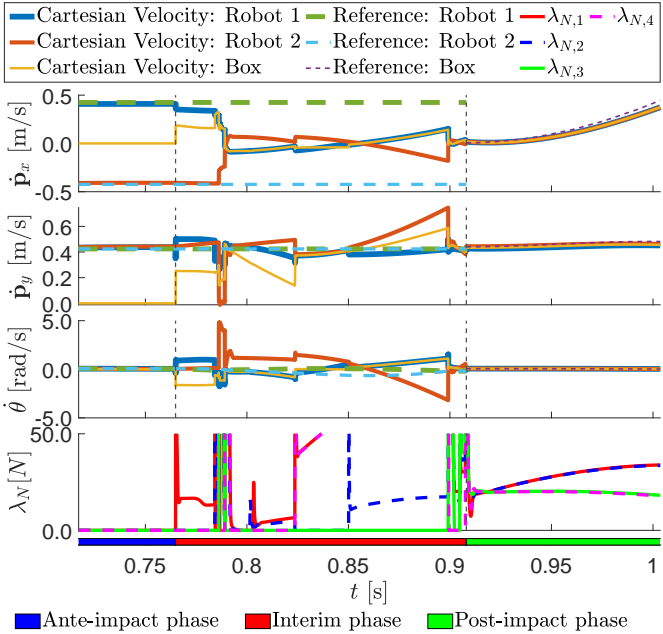}
		\caption{Cartesian end effector velocities and contact forces for the proposed control approach.}
		\label{fig:velocities_flex}
	\end{figure}

	In Figure \ref{fig:velocities_flex}, the velocities and contact forces around the impact sequence are depicted for simulations performed using the proposed control approach.  
	First of all, it can be seen that in the ante-impact phase, the velocity reference is followed for both robots. From the contact force profile, it can be observed that both end effectors reach the box in quick succession, with the main cause for the small separation of initial impact times being the uncertainty of the box pose.  
	This near-simultaneous impact with the desired velocity indicates that the control objectives from the ante-impact phase are met. 
	As soon as the first impact occurs, the system switches to the interim phase. During this phase, a series of unplanned intermediate impacts takes place as can be observed from the peaks in the contact forces and the many velocity jumps, which results from the misalignment between the impacting surfaces due to the uncertainty of the box pose in combination with the unmodeled robot flexibility. These velocity jumps and the inability to estimate the contact state in real applications confirm that velocity error feedback cannot reliably be used during this interim phase. 
	Once sustained contact is established and the post-impact phase is activated, it can be observed that, despite the unplanned intermediate impacts, the velocity of the box nearly matches the box velocity reference as a result of the post-impact velocity reference design that uses the predicted post-impact velocity. During this phase, the box does indeed follow the prescribed velocity reference, as well as the force distribution task, despite not using the real box state in the post-impact QP, further validating the approach.
	
	\begin{figure}
		\centering
		\includegraphics[width=0.98\linewidth]{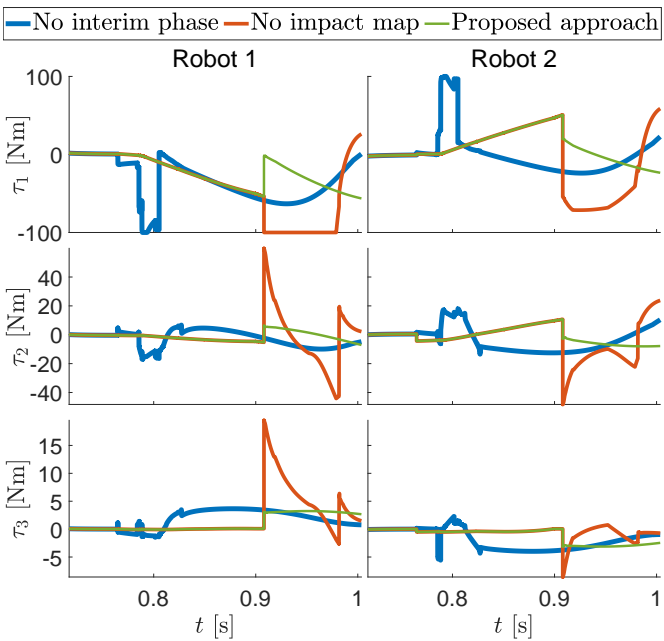}
		\caption{Input torques for the different control approaches.}
		\label{fig:torque_flex}
	\end{figure}
	
	Looking at the commanded torques for the proposed control approach against the two baseline approaches in Figure \ref{fig:torque_flex}, the benefit of the proposed control approach is further emphasized. In the approach with no interim phase, staying in the ante-impact phase until full contact is established results in an unreliable velocity signal to be used for velocity error feedback. This leads to large jumps in the input torque, potentially inducing additional vibrations. Hence, the approach is inferior to the proposed approach, where no velocity feedback is used, and only a subtle increase of the input torque over time can be observed as a result of the position feedback using the integrated ante-impact velocity reference. 
	When comparing to the proposed approach, it also becomes clear that using the approach with no impact map can result in a sizeable jump in the input torques after the impact, as there is a large discrepancy between the post-impact reference velocity of the box and its actual velocity, even resulting in the torque limits to be reached for joint 1. This sudden large jump can also trigger undesired vibrations, and is to be avoided. 
	
	A jump in the desired torque can be noticed also at the beginning of the post-impact phase for the desired approach. This is caused by the position feedback contribution in the interim mode error formulation of \eqref{eq:pos_task_int} that increases over time for the duration of the interim mode.  The reason for this is that $\bm p_{i,d}^\text{int}$ diverges from the end effector position $\bm p_i$ for both robots when they are in contact, while this position feedback term disappears in the post-impact mode. However, this jump is significantly smaller in magnitude when compared to the impact jump seen in the approach with no impact map, and due to the nature of the jump, it will cause a decrease in the magnitude of the input rather than an increase. Also, contrary to the approach with no impact map, the magnitude of the jump in the proposed approach will shrink when the impact event duration decreases, vanishing completely when the event is instantaneous. 
		
	
	\section{Conclusion}\label{sec:conclusion}
	
	In this work, a QP-based control approach is presented for swift impact-aware dual arm manipulation illustrated on a planar simulated case study. This approach extends the framework of time-invariant reference spreading, using a novel simulation-based approach to couple ante- and post-impact velocity reference fields. These fields are extended around the predicted impact locations, avoiding sudden large velocity errors and corresponding input torque jumps during and after an impact sequence. Two other novelties of the approach include synchronization during the ante-impact phase, which pursues a simultaneous impact between both end effectors and the object, and the post-impact phase design, pursuing contact force regulation and object manipulation without measurements of the object. 
	Simulation results using a realistic model of a torque controlled robot with flexible joints are used to validate the approach. These results show that unwanted input torque steps and spikes are minimized compared to two baseline approaches, highlighting the need for the interim-impact phase and references coupled by the impact dynamics. Future work involves scaling up to a full 3D case and performing real-life impact experiments. 

	\begin{ack}
	This work was partially supported by the Research Project I.AM. through the European Union H2020 program under GA 871899.
	\end{ack}
	
	\bibliography{library2}
	                                                   
	
	
	
	
	
	
	
	\appendix
	\section{Post-impact velocity estimation}\label{app:path_post}    
	The simulations used to create the post-impact box velocity estimation $\dot{\bm p}^+_{b,\text{est}}$ and $\dot{\theta}^+_{b,\text{est}}$ use the QP controller as described in Section \ref{sec:control_ante} and the dynamic model described in Section \ref{sec:dynamics}. 
	The system is initialized in multiple impact configurations where the robots simultaneously hit the box at different heights, indicated by positions $\bm p^-_{i,j}$ for simulation index $j \in \{1, 2, \dots, N_\text{exp}\}$ and robot index $i \in \{1,2\}$. The initial linear velocities $\dot{\bm p}^-_{i,j}$ are given by the desired velocity at positions $\bm p^-_{i,j}$, determined using \eqref{eq:lin_ref_ante}, and  $\dot{\theta}_{i,j} = 0$. 
	After simulating for a short given amount of time after which contacts between the box and the robots are fully established,  
	the velocity of the box is saved, defined as $\dot{\bm q}^+_{b,j} = [\dot{\bm p}^+_{b,j},\dot{\theta}^+_{b,j}]$. For each experiment $j$, the vertical offsets of both end effectors with respect to the box are defined as $\bm y^-_{j} = [y^-_{j,1}, y^-_{j,2}]$, with $y^-_{j,i}$ as the vertical translation of frame $o_ix_iy_i$ relative to frame $o_bx_by_b$, expressed in terms of $o_bx_by_b$. Using radial basis function (RBF) interpolation, a weighting matrix $\bm W = (\bm w_1, \bm w_2, \dots , \bm w_{N_\text{exp}}) \in \mathbb{R}^{3 \cross N_\text{exp}}$ is determined through
	\begin{equation}
	\bm W = \bm \Phi^{-1} \begin{bmatrix}
	\dot{\bm q}^+_{b,1} & \dot{\bm q}^+_{b,2} & \dots & \dot{\bm q}^+_{b,N_\text{exp}}
	\end{bmatrix},
	\end{equation}
	where $$
	\bm \Phi= \left[\begin{array}{cccc}
	\phi\left(\left\|\bm y^-_1-\bm y^-_1\right\|\right) & \cdots & \phi\left(\left\|\bm y^-_1-\bm y^-_{N_\text{exp}}\right\|\right) \\
	\vdots & \ddots & \vdots \\
	\phi\left(\left\|\bm y^-_{N_\text{exp}}-\bm y^-_1\right\|\right) & \cdots & \phi\left(\left\|\bm y^-_{N_\text{exp}}-\bm y^-_{N_\text{exp}}\right\|\right)
	\end{array}\right]$$
	with exponential radial basis function $\phi:\mathbb{R}\to\mathbb{R}$ as $$\phi(r)=e^{-(\rho r)^{2}}$$
	and user-defined shaping parameter $\rho \in \mathbb{R}^+$. The post-impact box velocity is obtained via RBF interpolation as 
	\begin{equation}
	\begin{bmatrix}\dot{\bm p}^+_{b,\text{est}} \\ \dot{\theta}^+_{b,\text{est}}\end{bmatrix} = \sum_{j=1}^{N_\text{exp}} \bm w_{j} \phi\left(\left\|\bm y^--\bm{y}^-_{j}\right\|\right)
	\end{equation}
	with $\bm y^-$ as the relative vertical displacement of the end effectors with respect to the estimated box pose at the beginning of the post-impact phase. 
\end{document}